%% file: main.tex
\documentclass{article}
\PassOptionsToPackage{numbers, compress}{natbib}
\usepackage[preprint]{neurips_2023}

\usepackage[utf8]{inputenc} 
\usepackage[T1]{fontenc}    
\usepackage[pagebackref=true,breaklinks=true,bookmarks=false,colorlinks=true]{hyperref}      
\usepackage{url}            
\usepackage{booktabs}       
\usepackage{amsfonts}       
\usepackage{nicefrac}       
\usepackage{microtype}      
\usepackage[table,xcdraw,dvipsnames]{xcolor}         
\usepackage{graphicx}
\usepackage{tikz}
\usetikzlibrary{positioning}
\usepackage{subcaption}
\usepackage{microtype}
\usepackage[normalem]{ulem}
\usepackage{adjustbox}
\usepackage{multirow}
\usepackage{tabularx}
\usepackage{makecell}
\usepackage{etoc}
\usepackage[accsupp]{axessibility}
\usepackage{xspace}
\usepackage{diagbox}
\usepackage{wrapfig}
\usepackage{bbding}
\usepackage{pifont}
\usepackage{wasysym}

\usepackage{caption}

\usepackage{CJKutf8}
\usepackage{svg}
\usepackage{graphicx}

\input{math_commands}

\usepackage{url}
\usepackage{cleveref} 
\Crefname{section}{Sec.}{Secs.}
\Crefname{table}{Tab.}{Tabs.}
\Crefname{figure}{Fig.}{Figs.}
\Crefname{equation}{Eq.}{Eqs.}
\Crefname{appendix}{Apx.}{Apx.}

\definecolor{citecolor}{HTML}{0071BC}
\hypersetup{colorlinks=true,citecolor=RoyalBlue}

\title{\Large Efficient Multimodal Learning from Data-centric Perspective} 

\author{
Muyang He$^{1,2\star}$
\quad Yexin Liu$^{1,3\star}$
\quad Boya Wu$^{1\star}$
\\
\textbf{
Jianhao Yuan$^{4}$
\quad Yueze Wang$^{1}$
\quad Tiejun Huang$^{1,2}$ 
\quad Bo Zhao$^{1,5\dagger}$  }
\\
$^1$Beijing Academy of Artificial Intelligence\\
$^2$Peking University \quad $^3$HKUST(GZ) \quad $^4$University of Oxford \quad $^5$ SJTU\\
$^\star$Equal Contribution \quad $^{\dagger}$Corresponding to Bo Zhao \texttt{$<$bo.zhao@sjtu.edu.cn$>$}\\
}

\begin{document}
\maketitle

\begin{abstract}
Multimodal Large Language Models (MLLMs) have demonstrated notable capabilities in general visual understanding and reasoning tasks. However, their deployment is hindered by substantial computational costs in both training and inference, limiting accessibility to the broader research and user communities. A straightforward solution is to leverage smaller pre-trained vision and language models, which inevitably cause significant performance drops. In this paper, we demonstrate the possibility of training a smaller but better MLLM with high-quality training data. Specifically, we introduce \emph{Bunny}, a family of lightweight MLLMs with flexible vision and language backbones for efficient multimodal learning from selected training data. Experiments show that our Bunny-4B/8B outperforms the state-of-the-art large MLLMs on multiple benchmarks. We expect that this work can provide the community with a clean and flexible open-source tool for further research and development. The code, models, and data can be found in \url{https://github.com/BAAI-DCAI/Bunny}.

\end{abstract}

\section{Introduction}
Recent advancements in Multimodal Large Language Models (MLLMs) have demonstrated exceptional visual understanding and reasoning performances across a range of tasks such as visual question answering \cite{team2023gemini, alayrac2022flamingo} and referring comprehension \cite{achiam2023gpt, peng2023kosmos, chen2023shikra}. Benefiting from scalable Transformer-based architecture \cite{vaswani2017attention} and web-scale training data sources, these models have become foundational in the field of artificial intelligence, with their parameters increasing from billions to trillions \cite{achiam2023gpt, touvron2023llama2, chowdhery2023palm}.

However, the deployment of these models is often hindered by their substantial computational costs and memory consumption in both training and inference phases, which limits their popularization across the broader research and user communities. Several early attempts, such as LLaVA-Phi~\cite{zhu2024llava}, Imp~\cite{imp2024}, and Vary-toy~\cite{wei2024small}, leverage off-the-shelf lightweight LLM backbones (e.g., Phi \cite{li2023textbooks, microsoft2024phi2}) and small vision encoders (e.g., SigLIP \cite{zhai2023sigmoid}) to build lightweight MLLMs. Further works also explore optimizations in model architecture and training recipes for tailored lightweight MLLM designs \cite{chu2023mobilevlm, minicpm2024}. 

While these attempts to create lightweight versions of MLLMs achieve various levels of success, these models often fall short in terms of performance when compared to their larger counterparts, as reducing the size of these models often leads to a compromise in model capacity. An alternative direction to bridge the performance gap is data optimization. Many works have shown the importance of high-quality training data. For instance, Phi-2 \cite{li2023textbooks} demonstrates that curated high-quality text data can bring the performances of small-scale LLMs close to that of large-scale models. SVIT \cite{zhao2023svit} validates the importance of data sources in MLLMs. In this work, we focus on data optimization to compensate for the reduction in model size.

We introduce \emph{Bunny}, a family of lightweight but powerful multimodal models with plug-and-play language/vision backbones and a multimodal projector. To compensate for the performance decrease caused by the model shrinking, we construct more informative training data by dataset condensation, i.e., curated data selection from a broader source. We demonstrate the possibility of training a smaller but better MLLM on more informative training data.
Extensive experiments indicate that our Bunny-4B/8B model outperforms not only state-of-the-art small MLLMs with similar sizes but also even large ones on popular benchmarks.

\section{Related Work}
\subsection{Large Language Model}
Recent advancements in Large Language Model (LLM) such as GPT \cite{radford2019language, brown2020language, achiam2023gpt}, LLaMA \cite{touvron2023llama,touvron2023llama2,llama3modelcard} and PaLM \cite{chowdhery2023palm, anil2023palm} have demonstrated remarkable capacity in various natural language processing tasks, for example, dialogue, creative writing, and problem-solving. Benefiting from scalable Transformer-based architecture \cite{vaswani2017attention} design and web-scale training data source, these models have become foundation models for general reasoning tasks. Among many research endeavors, one line of work focuses on miniaturizing LLM, for more efficient and low-cost training and inference. By leveraging high-quality pre-training and instruction tuning data, along with optimized lightweight architecture, lightweight LLMs with fewer than 3 billion parameters, such as Phi \cite{li2023textbooks, microsoft2024phi2}, StableLM-2 \cite{stablelm2023}, and TinyLLaMA \cite{zhang2024tinyllama}, have demonstrated performance comparable to larger models with 7/13 billion parameters.

\subsection{Multimodal Large Language Model}
Following the success of LLM, one line of research recently shifted further toward the Multimodal Large Language Model (MLLM) \cite{achiam2023gpt, team2023gemini} for unified cross-modality understanding and reasoning. 
Building upon various pre-trained LLM backbones, various methods are proposed to perform cross-modality fusion. Flamingo \cite{alayrac2022flamingo} and BLIP-2 \cite{li2023blip} propose different modality fusion techniques to fuse visual tokens to frozen large language models through gated attention or query transformers. Inspired by the success of instruction tuning, LLaVA \cite{liu2023llava, liu2023improvedllava,liu2024llavanext,li2024llavanext-strong} and MiniGPT-4 \cite{zhu2023minigpt, chen2023minigptv2} use visual instruction tuning to align visual input to LLMs, demonstrating remarkable success. More recent work, such as Kosmos-2 \cite{peng2023kosmos} and Shikra \cite{chen2023shikra}, further empowers MLLM with grounded visual understanding capacity. While these models exhibit promising potential for general-purpose visual reasoning and planning tasks, they are, in general, extremely expensive and prohibitive to train and deploy.

\subsection{Lightweight Multimodal Large Language Model}
More related to our work are those efforts dedicated to miniaturizing MLLMs, thereby achieving low-cost and fast deployment. LLaVA-Phi~\cite{zhu2024llava}, Imp~\cite{imp2024} and Vary-toy~\cite{wei2024small} leverage off-the-shelf small language models and also achieve comparable performance on various benchmarks. Several works explore various ways to optimize the model architecture and training recipe to compensate for the decrease in model size. For instance, MobileVLM~\cite{chu2023mobilevlm} focuses on the architecture design of lightweight LLMs and cross-modality projectors to enable training and inference on resource-limited mobile devices. Moreover, MiniCPM~\cite{minicpm2024} and TinyGPT-V~\cite{yuan2023tinygpt} explore Bayesian search on hyperparameters and complex multi-stage training strategy, respectively, to optimize training recipes. An alternative direction, yet under-explored, is training data optimization. A number of works have emphasized the critical role of training data in foundation models~\cite{chu2023mobilevlm, zhao2023svit}. In this work, without bells and whistles, we show that by optimizing the training data, small-scale models can readily outperform their large-scale counterparts.


\section{Bunny: A Family of Lightweight Multimodal Models}
This section introduces the flexible backbone combination of Bunny, the construction of informative training data, and the training procedure.

\subsection{Architecture and Backbone}
\begin{figure}
    \centering
    \includegraphics[width=0.9\linewidth]{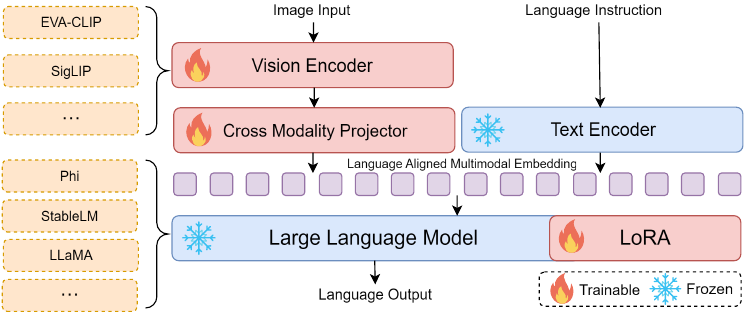}
    \caption{\textbf{Model Architecture.} Bunny offers a flexible choice of vision encoder and LLM backbone combination, which are aligned through the cross-modality projector.}
    \label{fig:arch}
\end{figure}

Following the success of the MLLM paradigm of leveraging visual instruction tuning to align vision and language representations, our model contains three main modules: the LLM backbone, Vision Encoder, and Cross-modality Projector. As illustrated in \Cref{fig:arch}, Bunny has a flexible combination of the three modules. We introduce some lightweight alternatives for each module as follows.

\paragraph{Large Language Model Backbone.}
We provide various options of state-of-the-art lightweight LLM: Phi-1.5 (1.3B) \cite{li2023textbooks}, Qwen1.5-1.8B \cite{qwen}, StableLM-2 (1.6B) \cite{stablelm2023}, MiniCPM-2B (2.7B) \cite{minicpm2024}, Phi-2 (2.7B) \cite{microsoft2024phi2}, Phi-3-Mini (3.8B) \cite{abdin2024phi} and Llama-3-8B \cite{llama3modelcard} as the language model backbone.

\paragraph{Vision Encoder.}
We provide two options for the lightweight vision encoder: SigLIP \cite{zhai2023sigmoid} and EVA-CLIP \cite{sun2023eva}, which are both efficient language-aligned image encoders with 428M parameters. 

\paragraph{Cross-modality Projector.} 
Following LLaVA \cite{liu2023improvedllava}, we leverage a two-layer MLP with a GELU activation function as a cross-modality projector to align the vision encoder and LLM. 

More choices of LLMs, vision encoders, and projectors will be provided in the future.

\subsection{Training Data Construction}
{The quality and scale of training data play critical roles in advancing MLLM performance. 
We construct \emph{Bunny-pretrain-LAION-2M} and \emph{Bunny-695K} for pre-training and instruction tuning, respectively.}


\paragraph{Pre-training Data.} We construct a high-quality pre-training data from LAION-2B \cite{schuhmann2022laion}, i.e., we condense LAION-2B into a 2M coreset for data-efficient learning. Specifically, we design a three-step coreset selection scheme\footnote{https://github.com/BAAI-DCAI/Dataset-Pruning/tree/main/LAION} based on CLIP embedding. Firstly, motivated by SemDeDup \cite{abbas2023semdedup}, we cluster all 2B image embeddings by k-means, and then in each cluster, build an undirected graph that any two embeddings are connected if their cosine similarity is above a predetermined threshold. Only one sample is kept in each connected sub-graph whose Euclidean distance to the cluster centroid ranks at the median. By setting the threshold to $0.86$, we obtain a subset of 952M. 
Secondly, we sort the remaining samples by the cosine similarity between their text embedding and image embedding and keep samples ranking $40\%-60\%$, subsequently resulting in a subset of 190M. In this way, the low-quality image-text pairs are removed.
Thirdly, we sort the remaining samples by the cosine similarity between its image embedding and its cluster centroid and keep samples ranking $15\%-35\%$, resulting in a subset of 38M, capturing the essence and diversity of LAION-2B. We finally randomly sampled 2 million samples from the 38M corset, resulting in Bunny-pretrain-LAION-2M for an appropriate training cost.

\paragraph{Fine-tuning Data.} We collect a set of visual instruction tuning datasets -- \emph{DataOptim}\footnote{https://github.com/BAAI-DCAI/DataOptim}, based on which we explore the better combination of fine-tuning datasets. Specifically, we leverage SVIT-mix-665K \cite{zhao2023svit} and replace ShareGPT-40K \cite{liu2023improvedllava} in it with WizardLM-evol-instruct-70K \cite{xu2023wizardlm}, resulting to Bunny-695K. We find that tuning MLLM on multimodal data may damage its cognition ability inherited from the pre-trained LLM. Probably, it is caused by the less informative and less diverse text in the multimodal training data. Keeping several high-quality pure text data in the fine-tuning dataset can relieve this problem. In experiments, we jointly use Bunny-695K, LLaVA-665K \cite{liu2023improvedllava} and ALLaVA-Instruct-4V \cite {chen2024allava} in the fine-tuning stage for better performance.

\subsection{Training Procedure}
We employ a two-stage training strategy. The first stage is the pre-training stage, where we align the visual embeddings from a pre-trained vision encoder with the text embeddings from the LLM. The second stage is the fine-tuning stage, where we apply visual instruction tuning to fully utilize the capabilities of the MLLM across various multimodal tasks. In both stages, we use the same cross-entropy loss for next-token prediction.
During the pre-training stage, only the cross-modality projector is optimized for one epoch. In the fine-tuning stage, we generally use LoRA \cite{hu2021lora} to train the LLM backbone, and the vision encoder and the cross-modality projector are fully tuned. In \cref{sec:abl}, we conduct ablation studies about LoRA, data, epoch and freezing component, and we also explore high image resolution and weight merging for more practical MLLMs. For more training details, please refer to our GitHub repository.


\section{Experiment}

\subsection{Settings}
We evaluate Bunny on eleven popular benchmarks: 
MME perception \cite{mme}, MME cognition \cite{mme}, MMBench \cite{mmbench} test split, MMBench \cite{mmbench} dev split, SEED-Bench-1 \cite{seedbench}, MMMU \cite{yue2023mmmu} validation split,  MMMU \cite{yue2023mmmu} test split, VQA-v2 \cite{goyal2017making} test-dev split, GQA \cite{GQA2019} test-dev-balanced split, ScienceQA-IMG \cite{lu2022learn} test split and POPE \cite{li-etal-2023-evaluating}: the averaged F1-score of three categories (random, popular and adversarial) on the validation set of MSCOCO,
to throughout assess its multimodal capacity. We compare against state-of-the-art multimodal large language models (MLLMs), including GPT-4V \cite{gpt-4}, BLIP-2 \cite{li2023blip}, InstructBLIP \cite{instructblip}, Shikra-13/7B \cite{chen2023shikra}, SVIT-v1.5-13B \cite{zhao2023svit}, LVIS-INSTRUCT4V-13B \cite{wang2023instruct4v}, ShareGPT4V-13B \cite{chen2023sharegpt4v}, VILA1.5-13/3B \cite{lin2023vila}, Mini-Gemini-HD-13B \cite{li2024mgm}, LLaVA-v1.5-13/7B \cite{liu2023improvedllava}, LLaVA-NeXT-13/7B \cite{liu2024llavanext}, MiniGPT-v2 \cite{chen2023minigptv2}, mPLUG-Owl2 \cite{ye2023mplugowl2}, SPHINX-Intern2 \cite{gao2024sphinx}, Yi-VL-6B \cite{ai2024yi}, DeepSeek-VL-7B \cite{lu2024deepseekvl}, MM1-7/3B-Chat \cite{mckinzie2024mm1}, Idefics2 \cite{laurenccon2024matters}, LLaVA-NeXT-Llama-3-8B \cite{li2024llavanext-strong}, MobileVLM-V2-3B \cite{chu2024mobilevlm}, TinyLLaVA-Phi-2-SigLIP-3.1B \cite{tinyllava}, Mipha-3B \cite{zhu2024comprehensive}, MiniCPM-V 2.0 \cite{minicpmv2}, Imp-v1.5-4B-Phi3 \cite{imp2024}.

\subsection{Comparison to the State of the Art}
\input{tables/main_result_v2}

As shown in \Cref{tab:main}, Bunny-4B (SigLIP-SO + Phi-3-Mini) and Bunny-8B (SigLIP-SO + Llama-3-8B) outperform MLLMs with similar sizes on most of the benchmarks, demonstrating exceptional multimodal understanding and reasoning capabilities. Remarkably, Bunny-8B achieves an 1644.1 score on MME$^\text{P}$, which overwhelms the runner-up LLaVA-NeXT-Llama-3-8B (1603.7) by a large margin. Bunny-4B also shows superiority over MM1-3B-Chat on the challenging MMMU benchmark, namely 7.5 and 4.7 improvements on the validation and test splits respectively. Besides, Bunny-4B and Bunny-8B accept images up to \textbf{1152$\times$1152.}


\subsection{Ablation Study}
\label{sec:abl}
\input{tables/ablation_lora}
\paragraph{LoRA vs. Fully Fine-tuning.} In the fine-tuning stage, we observe that LoRA empirically leads to better performance compared to fully tuning the LLM. This improvement is likely because smaller models are more prone to catastrophic forgetting, which LoRA tuning helps to mitigate. We show the experimental results in \Cref{tab:abl_lora}.

For below experiments, the LLM is always tuned using LoRA.

\input{tables/ablation_data}
\input{tables/ablation_data_2}
\paragraph{Fine-tuning Data.} \Cref{tab:abl_data} shows the ablation study on data strategy. Jointly leveraging Bunny-695K and LLaVA-665K for one epoch generally yields performance improvements across multiple benchmarks. A comparison between training on Bunny-695K for two epochs and training on Bunny-695K and LLaVA-665K for one epoch demonstrates that increasing data diversity enhances performance under the same training cost. To further enhance diversity, we incorporate ALLaVA-Instruct-4V \cite{chen2024allava}. Given the detailed answers provided in ALLaVA-Instruct-4V, we append ``Please answer the question in detail.'' to the end of questions to avoid overfitting the model to long-form answers. As shown in \Cref{tab:abl_data2}, adding ALLaVA-Instruct-4V brings performance improvements. Based on above results, we set the fine-tuning stage to be 1 epoch in the rest experiments in this paper.

\input{tables/ablation_vision}
\input{tables/ablation_vision_2}
\paragraph{Trainable vs. Frozen Vision Encoder.} By default, we freeze the vision encoder in both stages of our primary experiments. 
We also investigate the influence of enabling the vision encoder to be trainable during the fine-tuning stage. As shown in \Cref{tab:abl_vision}, the results vary depending on the fine-tuning data used. When employing Bunny-695K + LLaVA-665K, unfreezing the vision encoder leads to better results.
\Cref{tab:abl_vision2} shows that the performance can be further enhanced with additional finetuning data when the vision encoder is trainable.

\input{tables/ablation_pretrain_lr}
\input{tables/ablation_sft_lr}

\paragraph{Learning Rate.} As shown in \Cref{tab:abl_pre_lr}, increasing the pre-training learning rate does not consistently lead to improvements. In this context, both $5\times 10^{-4}$ and $8\times 10^{-4}$ are relatively good choices, while $1\times 10^{-3}$ is sub-optimal. As shown in \Cref{tab:abl_sft_lr}, the original learning rates set in LLaVA are the most appropriate. A learning rate that is too high causes extremely inferior results, including several sudden spikes in loss during training. 

\input{tables/ablation_res_pos}
\input{tables/ablation_res_s2}
\paragraph{Higher Resolution.} The ability of understanding high-resolution images and capturing the fine-grained details is crucial for a helpful MLLM.
Given that the default image size of SigLIP-SO is 384$\times$384, we try two approaches to scale up image resolution: positional embedding interpolation and sub-image decomposition. Through positional embedding interpolation, the vision encoder is adapted to the new resolution during fine-tuning, i.e. 448$\times$448, and an image is represented with 1024 tokens, resulting a 20\% training time increase. \Cref{tab:abl_res_pos} shows that the improvement is marginal. For sub-image decomposition, we leverage S$^2$-Wrapper \cite{shi2024we}, a simple mechanism that extends pretrained vision model to multiple image scales, which processes images at multiple scales by interpolating, splitting, encoding, merging, and concatenating features, resulting in a multi-scale representation with 3 times channel dimension while maintaining the token length.
It would increase the training time for 23\%. As shown in \Cref{tab:abl_res_s2}, S$^2$-Wrapper leads to a huge performance improvement while using Bunny-695K + LLaVA-665K. Considering the performance and high resolution, we take S$^2$-Wrapper as the final solution. 

\input{tables/ablation_merge}
\paragraph{Weight Merging.} During ablation studies, we obtain numerous models with the same architecture under different settings. We explore the potential benefits of merging two models by weighted averaging. As shown in \Cref{tab:abl_merge}, simple linear merging greatly enhances performance, with the best results forming our Bunny-4B/8B models. This method is more efficient and effective than hyper-parameter searching for improving model performance. We also observe that the results on some benchmarks (e.g. MME) are very sensitive to the weight factor.

\input{tables/ablation1}
\paragraph{Backbone Combination.} We also investigate Bunny's performance with various combinations of vision encoders and language models. As shown in \Cref{tab:abl1}, all combinations show exceptional performance, with SigLIP-SO \cite{zhai2023sigmoid} and Llama-3-8B \cite{llama3modelcard} achieving the best results. This highlights the flexibility of the Bunny framework and the high quality of the curated training data. Note that in the fine-tuning stage, Phi-1.5, StableLM-2 and Phi-2-based Bunny are trained on Bunny-695K for 1 epoch with the vision encoder frozen. In contrast, Phi-3-Mini and Llama-3-8B-based Bunny are trained using additional fine-tuning data, S$^2$-Wrapper, trainable vision encoder, weight merging, etc., as illustrated before.

\subsection{Qualitative Demonstrations}
We also present several qualitative test cases for Bunny. As illustrated in \Cref{fig:demo}, Bunny-8B with SigLIP-SO \cite{zhai2023sigmoid} and Llama-3-8B \cite{llama3modelcard} demonstrates exceptional capabilities in visual understanding, common sense reasoning, external knowledge referring, optical character recognizing, and mathematical problem-solving. These results feature Bunny-8B as a powerful and versatile visual assistant.
Additionally, \Cref{fig:demo_zh} presents several test cases in Chinese using Bunny-8B. Although we do not intentionally include Chinese in the multimodal training data, Bunny still exhibits impressive Chinese instruction-following capabilities. 

\input{figures/demo_table}

\input{figures/demo_table_zh}

\section{Conclusion}
We present Bunny, a family of lightweight but powerful multimodal models. It offers multiple plug-and-play vision encoders, including EVA-CLIP and SigLIP, and language backbones, including Phi-1.5, Qwen1.5-1.8B, StableLM-2, MiniCPM-2B, Phi-2, Phi-3-Mini, and Llama-3-8B. To compensate for the decrease in model size, we construct more informative training data by selecting data from a broader source. Remarkably, our Bunny-4B/8B outperforms the state-of-the-art large MLLMs on multiple benchmarks.

\clearpage
\newpage
\bibliography{main}{}
\bibliographystyle{unsrtnat}


\end{document}

%% file: math_commands.tex
\usepackage{amsmath,amsfonts,bm}

\def\eqref#1{equation~\ref{#1}}

\def\1{\bm{1}}

\DeclareMathAlphabet{\mathsfit}{\encodingdefault}{\sfdefault}{m}{sl}
\SetMathAlphabet{\mathsfit}{bold}{\encodingdefault}{\sfdefault}{bx}{n}



%% file: tables/main_result_v2.tex
\begin{table}
    \centering
    \caption{{ Comparison to state-of-the-art MLLMs on 11 benchmarks.} Our model outperforms them in most of the settings. We evaluate these models on benchmarks: MME$^\text{P}$: MME perception \cite{mme}, MME$^\text{C}$: MME cognition \cite{mme}, MMB$^\text{T}$: MMBench \cite{mmbench} test split, MMB$^\text{D}$: MMBench \cite{mmbench} dev split, SEED: SEED-Bench-1 \cite{seedbench} with both total accuracy and image accuracy, MMMU$^\text{V}$: MMMU \cite{yue2023mmmu} validation split,  MMMU$^\text{T}$: MMMU \cite{yue2023mmmu} test split, VQA-v2 \cite{goyal2017making} test-dev split, GQA \cite{GQA2019} test-dev-balanced split, SQA$^\text{I}$: ScienceQA-IMG \cite{lu2022learn} test split and POPE \cite{li-etal-2023-evaluating}: the averaged F1-score of three categories (random, popular and adversarial) on the validation set of MSCOCO.
    We mark the best performance \textbf{bold} and the runner-up \underline{underlined} in each section.
    $^\mathsection$We evaluate the officially released checkpoint by ourselves.}
   \setlength{\tabcolsep}{3pt}
 \resizebox{\textwidth}{!}{%
        \begin{tabular}{l | ccccccccccc}
            \toprule
            Model  & MME$^\text{P}$ & MME$^\text{C}$ & MMB$^\text{T}$ & MMB$^\text{D}$ & SEED(-I) & MMMU$^\text{V}$ & MMMU$^\text{T}$ & VQA$^\text{v2}$ & GQA & SQA$^\text{I}$ & POPE \\
            \midrule


            \multicolumn{11}{l}{\emph{Size $>$ 8B}} \\
            \midrule

            GPT-4V-1106 \cite{gpt-4} & 1334.0 & 437.5 & 77.0 & 75.1 & --/71.6 & 53.8 & -- & -- & -- & 82.1 & 75.4 \\
            
            BLIP-2-Flan-T5-XXL \cite{li2023blip} & 1293.8 & 290.0 & -- & -- & --/-- & 35.4 & 34.0 & 65.0 & 44.6 & 64.5 & -- \\
            
            InstructBLIP-Flan-T5-XXL \cite{instructblip}  & 1212.8 & 291.8 & -- & -- & --/-- & 35.7 & 33.8 & -- & 47.9 & 70.6 & -- \\
            
            BLIP-2-Vicuna-13B \cite{li2023blip} & -- & -- & -- & -- & --/-- & -- & -- & -- & 41.0 & 61.0 & -- \\
            
            InstructBLIP-Vicuna-13B \cite{instructblip} & -- & -- & -- & -- & --/-- & --& -- & -- & 49.5 & 63.1 & 83.7 \\
            
            Shikra-13B \cite{chen2023shikra} & -- & -- & -- & -- & --/-- & -- & -- & 77.4 & -- & -- & -- \\

            SVIT-v1.5-13B \cite{zhao2023svit} & 1565.8 & 323.2 & 69.1 & -- & 61.9/-- & -- & 33.3 & 80.3 & 64.1 & 70.0 & -- \\ 

            LVIS-INSTRUCT4V-13B \cite{wang2023instruct4v} & 1574.9 & 286.8 & -- & 68.0 & 61.6/-- & -- & -- & 80.7 & 63.6 & 70.6 & 86.0 \\
            
            ShareGPT4V-13B \cite{chen2023sharegpt4v} & {1618.7} & 303.2 & -- &68.5 & --/70.8 &--& -- & 81.0 & {64.8} & 71.2 & -- \\


            VILA1.5-13B \cite{lin2023vila} & 1569.6 & -- & -- & 74.9 & 65.1/72.6 & 37.9 & 33.6 & 82.8 & 64.3 & 80.1 & 86.3 \\

            Mini-Gemini-HD-13B \cite{li2024mgm} & 1597.0 & 320.0 & -- & 68.6 & --/-- & 37.3 & 35.1 & -- & -- & -- & -- \\
            
            LLaVA-v1.5-13B (LoRA) \cite{liu2023improvedllava}  & 1541.7 & 300.4$^\mathsection$ & 
            68.4$^\mathsection$ & 68.5 & 61.3/-- & 40.0$^\mathsection$ & 33.2$^\mathsection$ & 80.0 & 63.3 & 71.2 & 86.7 \\

            LLaVA-NeXT-13B \cite{liu2024llavanext} & 1575.0 & 326.0 & -- & 70.0 & --/71.9 & 36.2 & -- & 82.8 & 65.4 & 73.6 & 86.2 \\
            
            \midrule  
            \multicolumn{11}{l}{\emph{4B $<$ Size $\leq$ 8B}} \\
            \midrule
            
            InstructBLIP-Vicuna-7B \cite{instructblip} & -- & -- & 33.9 & 36.0 & 53.4/--& -- & -- & -- & 49.2 & 60.5 & -- \\
            
            MiniGPT-v2 \cite{chen2023minigptv2} & -- & -- & -- & -- & --/-- & -- & -- & -- & 60.3 & -- & -- \\
            
            Shikra-7B \cite{chen2023shikra} & -- & -- & 60.2 & 58.8 & --/-- & -- & -- & -- & -- & -- & -- \\
            
            mPLUG-Owl2 \cite{ye2023mplugowl2} & 1450.2 & 313.2 & 66.0 & 66.5 & 57.8/-- & 32.7 & 32.1 & 79.4 & 56.1 & 68.7 & 85.8 \\

            
            SPHINX-Intern2 \cite{gao2024sphinx} & 1260.4 & 294.6 & -- & 57.9 & --/68.8 & -- & -- & 75.5 & 56.2 & 70.4  & 86.9 \\

            Yi-VL-6B \cite{ai2024yi} & -- & -- & 68.4 & 68.6 & --/67.6 & 39.1 & \underline{37.8} & -- & -- & --  & -- \\

            DeepSeek-VL-7B \cite{lu2024deepseekvl} & -- & -- & -- & \underline{73.2} & --/\underline{70.4} & 36.6  & -- & --  & -- & -- & \textbf{88.1} \\

            LLaVA-v1.5-7B (LoRA) \cite{liu2023improvedllava}  & 1476.9 & 267.9$^\mathsection$ &  66.1$^\mathsection$ & 66.1 & 60.1/-- & 34.4$^\mathsection$ & 31.7$^\mathsection$ & 79.1 & 63.0 & 68.4 & 86.4 \\

            LLaVA-NeXT-7B \cite{liu2024llavanext} & 1519.0 & {332.0} & -- & 67.4 & --/70.2 & 35.8 & -- & 81.8 & \textbf{64.2} & 70.1  & 86.5 \\

            MM1-7B-Chat \cite{mckinzie2024mm1} & 1529.3 & 328.9 & -- & 72.3 & \underline{64.0}/69.9 & 37.0 & 35.6 & \underline{82.8} & -- & \underline{72.6}  & 86.6 \\

            Idefics2 \cite{laurenccon2024matters} & -- & -- & \underline{76.7} & -- & --/-- & \underline{43.0} & {37.7} & 81.2 & -- & --  & -- \\

            LLaVA-NeXT-Llama-3-8B \cite{li2024llavanext-strong} & \underline{1603.7} & \textbf{367.8} & -- & 72.1 & --/-- & 41.7 & -- & -- & -- & -- & -- \\
            
            
            \textbf{Bunny-8B} & \textbf{1644.1} & \underline{367.5} & \textbf{78.1} & \textbf{77.2} & \textbf{66.2}/\textbf{73.5}& \textbf{43.3} & \textbf{39.0} & \textbf{82.9} & \underline{64.0} & \textbf{79.9} & \underline{87.2} \\

            \midrule  
            \multicolumn{11}{l}{\emph{Size $\leq$ 4B}} \\
            \midrule
            

            
            
            MobileVLM-V2-3B \cite{chu2024mobilevlm} & 1440.5 & -- & -- & 63.2 & --/-- & -- & -- & -- & 61.1 & 70.0 & 84.7 \\
            
            TinyLLaVA-Phi-2-SigLIP-3.1B \cite{tinyllava} & 1466.4 & -- & -- & -- & --/-- & \underline{38.4} & -- & 80.1 & 62.1 & 73.0 & \underline{87.2} \\

            VILA1.5-3B \cite{lin2023vila} & 1442.4 & -- & -- & 63.4 & 60.9/67.9 & 33.3 & 30.8 & 80.4 & 61.5 & 69.0 & 85.9 \\

            Mipha-3B \cite{zhu2024comprehensive} & 1488.9 & 295.0 & -- & 69.7 & --/\underline{68.9} & -- & -- & 81.3 & \textbf{63.9} & 70.9 & 86.7 \\

            MiniCPM-V 2.0 \cite{minicpmv2} & 1411.4 & \textbf{396.8} & \underline{69.1} & 69.6 & --/67.1 & {38.2} & -- & -- & -- & \textbf{80.7} & 86.3 \\

            MM1-3B-Chat \cite{mckinzie2024mm1} & 1482.5 & 279.3 & -- & 67.8 & \underline{63.0}/{68.8} & 33.9 & \underline{33.7} & \underline{82.0} & -- & 69.4  & \textbf{87.4} \\
            
            Imp-v1.5-4B-Phi3 \cite{imp2024} & \underline{1507.7} & -- & -- & \underline{73.3} & --/-- & -- & -- & 81.5 & \underline{63.5} & \underline{78.3} & 86.9 \\
            
            \textbf{Bunny-4B} & \textbf{1581.5} & \underline{361.1} & \textbf{75.7} & \textbf{74.2} & \textbf{64.9}/\textbf{72.5} & \textbf{41.4} & \textbf{38.4} & \textbf{82.1} & 63.2 & \underline{78.3} & \underline{87.2} \\
            

            \bottomrule
        \end{tabular}
    }
    \label{tab:main}
\end{table}

%% file: tables/ablation_lora.tex
\begin{table}
    \centering
\caption{Comparison between LoRA and fully tuning. The model is composed of SigLIP-SO \cite{zhai2023sigmoid} and Phi-2 \cite{microsoft2024phi2}, and trained on Bunny-695K for 1 epoch. The vision encoder is frozen.}
   \setlength{\tabcolsep}{3pt}

\resizebox{\textwidth}{!}{\begin{tabular}{l|ccccccccccc}
\toprule
{LLM Tuning} & {MME$^\text{P}$} & {MME$^\text{C}$} &  MMB$^\text{T}$ & MMB$^\text{D}$ & {SEED(-I)} & MMMU$^\text{V}$ & MMMU$^\text{T}$ & {VQA$^\text{v2}$} & {GQA} & {SQA$^\text{I}$} & POPE \\
\midrule
Fully Tuning  & 1465.2 & 277.5 & 67.9 & 67.2 & 62.1(69.5) & 36.8 & \textbf{33.1} & 79.1 & 61.7 & \textbf{72.8} & 85.9 \\
LoRA & \textbf{1488.8} & \textbf{289.3} & \textbf{69.2} & \textbf{68.6} & \textbf{62.5}(\textbf{70.6}) & \textbf{38.2} & 33.0 & \textbf{79.8} & \textbf{62.5} & 70.9 & \textbf{86.8} \\
\bottomrule
\end{tabular}}
    \label{tab:abl_lora}
\end{table}

%% file: tables/ablation_data.tex
\begin{table}
    \centering
\caption{Ablation study on fine-tuning data and training epochs. The model is composed of SigLIP-SO \cite{zhai2023sigmoid} and Phi-2 \cite{microsoft2024phi2} with the vision encoder frozen.}
   \setlength{\tabcolsep}{3pt}

\resizebox{\textwidth}{!}{\begin{tabular}{lc|ccccccccccc}
\toprule
{Fine-tuning Data} & Epoch &{MME$^\text{P}$} & {MME$^\text{C}$} &  MMB$^\text{T}$ & MMB$^\text{D}$ & {SEED(-I)} & MMMU$^\text{V}$ & MMMU$^\text{T}$ & {VQA$^\text{v2}$} & {GQA} & {SQA$^\text{I}$} & POPE \\
\midrule
Bunny-695K &1 & {1488.8} & \textbf{289.3} & {69.2} & {68.6} & {62.5(70.6)} & \textbf{38.2} & 33.0 & {79.8} & {62.5} & \textbf{70.9} & {86.8} \\
Bunny-695K &2 & 1480.6 & 272.5 & {71.2} & {69.0} & \textbf{63.0}(\textbf{71.3}) & 37.6 & 32.5 & \textbf{81.0} & \textbf{64.0} & 69.7 & 86.3 \\
Bunny-695K +  LLaVA-665K & 1 & \textbf{1501.3} & 272.5 & \textbf{72.4} & \textbf{71.1} & \textbf{63.0}(71.1) & 37.8 & \textbf{33.4} & \textbf{81.0} & {63.6} & 69.5 & \textbf{87.2} \\
\bottomrule
\end{tabular}}
    \label{tab:abl_data}
\end{table}

%% file: tables/ablation_data_2.tex
\begin{table}
    \centering
\caption{Ablation study on extra fine-tuning data. The model is composed of SigLIP-SO \cite{zhai2023sigmoid} and Llama-3-8B \cite{llama3modelcard} where the vision encoder is frozen. S$^2$-Wrapper \cite{shi2024we} is used.} 
   \setlength{\tabcolsep}{3pt}

\resizebox{\textwidth}{!}{\begin{tabular}{lc|ccccccccccc}
\toprule
{Fine-tuning Data} & Epoch &{MME$^\text{P}$} & {MME$^\text{C}$} &  MMB$^\text{T}$ & MMB$^\text{D}$ & {SEED(-I)} & MMMU$^\text{V}$ & MMMU$^\text{T}$ & {VQA$^\text{v2}$} & {GQA} & {SQA$^\text{I}$} & POPE \\
\midrule
 Bunny-695K  &1&\textbf{1614.7} & {310.0} & {75.6} & {75.4} & {65.2}({72.6})&\textbf{42.3} & {38.2} & {81.5} & {63.4} & \textbf{80.4} & \textbf{87.3} \\
 Bunny-695K + ALLaVA-Instruct-4V &1 & {1599.6} & \textbf{348.9} & \textbf{76.3} & \textbf{75.6} & \textbf{66.2}(\textbf{73.4}) & 41.6 & \textbf{38.3} & \textbf{81.9} & \textbf{64.1} & 79.5 & {86.4} \\
\bottomrule
\end{tabular}}
    \label{tab:abl_data2}
\end{table}

%% file: tables/ablation_vision.tex
\begin{table}
    \centering
\caption{Ablation study on tuning/freezing vision encoder in the fine-tuning stage. The model consists of SigLIP-SO \cite{zhai2023sigmoid} and Phi-2 \cite{microsoft2024phi2}.  
}
   \setlength{\tabcolsep}{3pt}

\resizebox{\textwidth}{!}{\begin{tabular}{lc|ccccccccccc}
\toprule
 Fine-tuning Data & Vision Encoder &{MME$^\text{P}$} & {MME$^\text{C}$} &  MMB$^\text{T}$ & MMB$^\text{D}$ & {SEED(-I)} & MMMU$^\text{V}$ & MMMU$^\text{T}$ & {VQA$^\text{v2}$} & {GQA} & {SQA$^\text{I}$} & POPE \\
\midrule
\multirow{2}{*}{Bunny-695K}  & Frozen & \textbf{1488.8} & \textbf{289.3} & {69.2} & {68.6} & \textbf{62.5}(70.6) & \textbf{38.2} & 33.0 & {79.8} & {62.5} & \textbf{70.9} & \textbf{86.8} \\ 
 & Trainable & {1486.0} & {281.1} & \textbf{69.5} & \textbf{69.3} & {62.3(70.6)} & {37.7} & \textbf{33.2} & \textbf{79.9} & \textbf{62.6} & 70.6 & {85.9} \\ 
\midrule
\multirow{2}{*}{Bunny-695K + LLaVA-665K}  & Frozen & {1501.3} & 272.5 & \textbf{72.4} & {71.1} & {63.0(71.1)} & 37.8 & \textbf{33.4} & {81.0} & {63.6} & 69.5 & \textbf{87.2} \\
& Trainable & \textbf{1521.5} & \textbf{291.8} & {71.9} & \textbf{71.3} & \textbf{63.1}(71.1) & \textbf{38.0} & {33.1} & \textbf{81.1} & \textbf{63.7} & \textbf{70.8} & {87.0} \\
\bottomrule
\end{tabular}}
    \label{tab:abl_vision}
\end{table}

%% file: tables/ablation_vision_2.tex
\begin{table}
    \centering
\caption{Ablation study on fine-tuning data with vision encoder tuned. The experiments are conducted with SigLIP-SO \cite{zhai2023sigmoid} as the vision encoder, and S$^2$-Wrapper \cite{shi2024we} is enabled. B-695K, L-665K and A-Ins represents for Bunny-695K, LLaVA-665K and ALLaVA-Instruct-4V, respectively.
}
   \setlength{\tabcolsep}{3pt}

\resizebox{\textwidth}{!}{\begin{tabular}{ll|ccccccccccc}
\toprule
LLM &{Fine-tuning Data}  &{MME$^\text{P}$} & {MME$^\text{C}$} &  MMB$^\text{T}$ & MMB$^\text{D}$ & {SEED(-I)} & MMMU$^\text{V}$ & MMMU$^\text{T}$ & {VQA$^\text{v2}$} & {GQA} & {SQA$^\text{I}$} & POPE \\
\midrule
\multirow{2}{*}{Phi-3-Mini} & B-695K + L-665K & {1503.5} & {336.4} & {74.4} & {74.3} & {64.2}({71.4})&{38.8} &\textbf{38.6} & {81.6} & \textbf{63.1} & {75.0} & \textbf{87.0} \\
 & B-695K + L-665K + A-Ins   &\textbf{1590.1} & \textbf{343.9} & \textbf{75.6} & \textbf{74.4} & \textbf{64.4}(\textbf{71.9})&\textbf{42.4} & 38.1 & \textbf{82.0} & {62.9} & \textbf{77.8} & {86.9} \\
\midrule
\multirow{2}{*}{Llama-3-8B} & B-695K + L-665K  &{1571.2} & {314.3} & {75.9} & {75.5} & {64.8}({72.6})&\textbf{42.2} & {37.6} & {82.4} & \textbf{64.8} & \textbf{79.2} & {86.8} \\
& B-695K + L-665K + A-Ins  &\textbf{1649.7} & \textbf{341.1} & \textbf{77.2} & \textbf{76.5} & \textbf{65.5}(\textbf{73.1})&{40.4} & \textbf{38.8} & \textbf{82.8} & {64.1} & {79.0} & \textbf{87.3} \\
\bottomrule
\end{tabular}}
    \label{tab:abl_vision2}
\end{table}

%% file: tables/ablation_pretrain_lr.tex
\begin{table}
    \centering
\caption{Ablation study on pre-training learning rate. The model consists of SigLIP-SO \cite{zhai2023sigmoid} and Phi-2 \cite{microsoft2024phi2}, where Bunny-695K is utilized for fine-tuning and the vision encoder is frozen.
}
   \setlength{\tabcolsep}{3pt}

\resizebox{\textwidth}{!}{\begin{tabular}{l|ccccccccccc}
\toprule
 Pre-training LR & {MME$^\text{P}$} & {MME$^\text{C}$} &  MMB$^\text{T}$ & MMB$^\text{D}$ & {SEED(-I)} & MMMU$^\text{V}$ & MMMU$^\text{T}$ & {VQA$^\text{v2}$} & {GQA} & {SQA$^\text{I}$} & POPE \\
\midrule
$5\times 10^{-4}$ & {1488.8} & \textbf{289.3} & {69.2} & {68.6} & \textbf{62.5}(\textbf{70.6}) & \textbf{38.2} & 33.0 & {79.8} & \textbf{62.5} & 70.9 & {86.8} \\
$8\times 10^{-4}$  & \textbf{1494.3} & 277.5 & 69.6 & \textbf{69.0} & 62.1(70.5) & 36.4 & \textbf{33.5} & \textbf{80.0} & 62.3 & \textbf{72.9} & \textbf{86.9} \\
$1\times 10^{-3}$  & 1448.3 & 261.1 & \textbf{70.5} & 68.8 & 62.4(70.4) & 36.8 & {33.3} & \textbf{80.0} & 62.3 & {71.1} & 86.4 \\
\bottomrule
\end{tabular}}
    \label{tab:abl_pre_lr}
\end{table}

%% file: tables/ablation_sft_lr.tex
\begin{table}
    \centering
\caption{Ablation study on fine-tuning learning rate. The model consists of SigLIP-SO \cite{zhai2023sigmoid} and Phi-2 \cite{microsoft2024phi2}, where Bunny-695K + LLaVA-665K is utilized for fine-tuning and the vision encoder is trainable.
}
   \setlength{\tabcolsep}{3pt}

\resizebox{\textwidth}{!}{\begin{tabular}{cc|ccccccccccc}
\toprule
 LoRA LR &  Vision \& Projector LR &{MME$^\text{P}$} & {MME$^\text{C}$} &  MMB$^\text{T}$ & MMB$^\text{D}$ & {SEED(-I)} & MMMU$^\text{V}$ & MMMU$^\text{T}$ & {VQA$^\text{v2}$} & {GQA} & {SQA$^\text{I}$} & POPE \\
\midrule
$2\times 10^{-4}$ &  $2\times 10^{-5}$& \textbf{1521.5} & \textbf{291.8} & \textbf{71.9} & \textbf{71.3} & \textbf{63.1}(\textbf{71.1}) & {38.0} & {33.1} & \textbf{81.1} & \textbf{63.7} & {70.8} & \textbf{87.0} \\
$1\times 10^{-4}$ & $1\times 10^{-5}$ & {1492.8} & 283.9 & 71.2 & 69.7 & \textbf{63.1}(70.9) & \textbf{38.4} & \textbf{33.3} & {80.4} & 62.8 & \textbf{71.5} & {86.8} \\
$4\times 10^{-4}$ & $4\times 10^{-5}$ & {987.7} & 240.4 & 0.0 & 0.0 & 25.9(26.3) & 27.9 & {25.3} & {69.0} & 53.7 & {36.7} & {83.3} \\
\bottomrule
\end{tabular}}
    \label{tab:abl_sft_lr}
\end{table}

%% file: tables/ablation_res_pos.tex
\begin{table}[t]
    \centering
\caption{Results on scaling up image resolution by positional embedding interpolation. The model consists of SigLIP-SO \cite{zhai2023sigmoid} and Phi-2 \cite{microsoft2024phi2}.}
   \setlength{\tabcolsep}{3pt}

\resizebox{\textwidth}{!}{\begin{tabular}{lcc|ccccccccccc}
\toprule
{Fine-tuning Data} & Vision Encoder & Res &{MME$^\text{P}$} & {MME$^\text{C}$} &  MMB$^\text{T}$ & MMB$^\text{D}$ & {SEED(-I)} & MMMU$^\text{V}$ & MMMU$^\text{T}$ & {VQA$^\text{v2}$} & {GQA} & {SQA$^\text{I}$} & POPE \\
\midrule
\multirow{2}{*}{Bunny-695K + LLaVA-665K} & \multirow{2}{*}{Trainable} & 384 & \textbf{1521.5} & {291.8} & \textbf{71.9} & {71.3} & {63.1(71.1)} & \textbf{38.0} & \textbf{33.1} & {81.1} & \textbf{63.7} & {70.8} & {87.0} \\
&&448 &{1489.2} & \textbf{309.6} & {71.6} & \textbf{71.5} & \textbf{63.3}(\textbf{71.4}) & {37.2} & {32.5} & \textbf{81.4} & {63.2} & \textbf{71.0} & \textbf{87.2} \\
\bottomrule
\end{tabular}}
    \label{tab:abl_res_pos}
\end{table}

%% file: tables/ablation_res_s2.tex
\begin{table}
    \centering
\caption{Ablation study on scaling up image resolution by sub-image decomposition (S$^2$-Wrapper \cite{shi2024we}). The model consists of SigLIP-SO \cite{zhai2023sigmoid} and Phi-3-Mini \cite{abdin2024phi}.}
   \setlength{\tabcolsep}{3pt}

\resizebox{\textwidth}{!}{\begin{tabular}{lcc|ccccccccccc}
\toprule
{Fine-tuning Data} & Vision Encoder & Res &{MME$^\text{P}$} & {MME$^\text{C}$} &  MMB$^\text{T}$ & MMB$^\text{D}$ & {SEED(-I)}  & MMMU$^\text{V}$ & MMMU$^\text{T}$ & {VQA$^\text{v2}$} & {GQA} & {SQA$^\text{I}$} & POPE \\
\midrule
\multirow{2}{*}{Bunny-695K} & \multirow{2}{*}{Frozen} & 384 &{1402.0} & {286.1} & \textbf{73.1} & \textbf{72.3} & \textbf{64.6}(\textbf{71.9})&{40.7} & \textbf{39.3} & {80.6} & {62.4} & {75.3} & {85.8} \\
& & 1152 & \textbf{1473.1} & \textbf{332.1} & {72.8} & {72.0} & {64.1}({71.3}) &\textbf{41.8} & {38.3} & {80.6} & {62.4} & \textbf{75.6} & \textbf{86.6} \\
\midrule
\multirow{2}{*}{Bunny-695K + LLaVA-665K } & \multirow{2}{*}{Trainable} & 384 &{1490.5} & {316.4} & {72.4} & {72.0} & {63.7}({71.3}) &\textbf{40.0} & {38.3} & {81.3} & {63.0} & {74.4} & {86.4} \\
 &  & 1152 &\textbf{1503.5} & \textbf{336.4} & \textbf{74.4} & \textbf{74.3} & \textbf{64.2}(\textbf{71.4}) &{38.8} & \textbf{38.6} & \textbf{81.6} & \textbf{63.1} & \textbf{75.0} & \textbf{87.0} \\
\bottomrule
\end{tabular}}
    \label{tab:abl_res_s2}
\end{table}

%% file: tables/ablation_merge.tex
\begin{table}
    \centering
\caption{{Ablation study on weight merging. The experiments are conducted with SigLIP-SO \cite{zhai2023sigmoid} as the vision encoder. The weight factor is selected empirically. \textbf{F}/\textbf{T} represents whether the \textbf{V}ision \textbf{E}ncoder is \textbf{F}rozen or \textbf{T}rainable. B-695K, L-665K and A-Ins represents for Bunny-695K, LLaVA-665K and ALLaVA-Instruct-4V, respectively. }}

   \setlength{\tabcolsep}{3pt}

\resizebox{\textwidth}{!}{\begin{tabular}{lclc|ccccccccccc}
\toprule
LLM  & S$^2$ &{Fine-tuning Data} & VE  &{MME$^\text{P}$} & {MME$^\text{C}$} &  MMB$^\text{T}$ & MMB$^\text{D}$ & {SEED(-I)}  & MMMU$^\text{V}$ & MMMU$^\text{T}$ & {VQA$^\text{v2}$} & {GQA} & {SQA$^\text{I}$} & POPE \\
\midrule
\multirow{3}{*}{Phi-3-Mini} & \multirow{3}{*}{w$\backslash$o} & B-695K &F  &{1402.0} & {286.1} & {73.1} & {72.3} & \textbf{64.6}({71.9})&\textbf{40.7} & \textbf{39.3} & {80.6} & {62.4} & \textbf{75.3} & {85.8} \\
 &  & B-695K + L-665K &T  &{1490.5} & {316.4} & {72.4} & {72.0} & {63.7}({71.3})&{40.0} & {38.3} & {81.3} & {63.0} & {74.4} & {86.4} \\
  &  & \multicolumn{2}{c|}{3:7 Averaging}  &\textbf{1495.2} & \textbf{338.9} & \textbf{74.0} & \textbf{73.5} & {64.5}(\textbf{72.1})&{40.1} & {39.1} & \textbf{81.5} & \textbf{63.5} & {75.2} & \textbf{86.7} \\
\midrule
\multirow{3}{*}{Phi-3-Mini} & \multirow{3}{*}{w$\backslash$} & B-695K &F  &{1473.1} & {332.1} & {72.8} & {72.0} & {64.1}({71.3})&\textbf{41.8} & {38.3} & {80.6} & {62.4} & {75.6} & {86.6} \\
 &  & B-695K + L-665K &T  &{1503.5} & {336.4} & \textbf{74.4} & \textbf{74.3} & {64.2}({71.4})&{38.8} & {38.6} & {81.6} & {63.1} & {75.0} & \textbf{87.0} \\
  &  & \multicolumn{2}{c|}{3:7 Averaging}  &\textbf{1503.9} & \textbf{362.9} & {74.1} & {74.1} & \textbf{64.6}(\textbf{71.7})&{40.2} & \textbf{38.8} & \textbf{81.7} & \textbf{63.4} & \textbf{76.3} & \textbf{87.0} \\
  \midrule
\multirow{3}{*}{Phi-3-Mini} & \multirow{3}{*}{w$\backslash$} & B-695K + A-Ins & F  & {1552.4} & 306.8 & {74.0} & {73.8} & {62.8}(71.0) & 40.6 & 38.1 & {81.0} & {62.3} & 75.0 & {86.7} \\
 &  & B-695K + L-665K + A-Ins & T  &\textbf{1590.1} & {343.9} & {75.6} & \textbf{74.4} & {64.4}({71.9})&\textbf{42.4} & 38.1 & {82.0} & {62.9} & {77.8} & {86.9} \\
  &  & \multicolumn{2}{c|}{3:7 Averaging (\textbf{Bunny-4B})}  &{1581.5} & \textbf{361.1} & \textbf{75.7} & {74.2} & \textbf{64.9}(\textbf{72.5}) & {41.4} & \textbf{38.4} & \textbf{82.1} & \textbf{63.2} & \textbf{78.3} & \textbf{87.2} \\
\midrule
  \multirow{3}{*}{Llama-3-8B} & \multirow{3}{*}{w$\backslash$o} & B-695K &F  &{1542.7} & \textbf{344.6} & {75.7} & {74.8} & {65.2}({73.1})&{42.4} & {37.1} & {82.0} & {64.4} & {78.7} & {85.9} \\
 &  & B-695K + L-665K &T  &{1562.2} & {319.6} & {75.3} & {74.4} & {65.2}({72.5})&\textbf{43.4} & {38.0} & {82.4} & {64.1} & {77.4} & {86.4} \\
  &  & \multicolumn{2}{c|}{5:5 Averaging}  &\textbf{1588.9} & {321.1} & \textbf{77.2} & \textbf{76.7} & \textbf{65.9}(\textbf{73.3})&{42.8} & \textbf{39.0} & \textbf{82.6} & \textbf{64.8} & \textbf{80.4} & \textbf{86.9} \\
  \midrule
    \multirow{3}{*}{Llama-3-8B} & \multirow{3}{*}{w$\backslash$} & B-695K &F  &\textbf{1614.7} & {310.0} & {75.6} & {75.4} & {65.2}({72.6})&{42.3} & {38.2} & {81.5} & {63.4} & \textbf{80.4} & {87.3} \\
 &  & B-695K + L-665K &T  &{1571.2} & {314.3} & {75.9} & {75.5} & {64.8}({72.6})&{42.2} & {37.6} & \textbf{82.4} & \textbf{64.8} & {79.2} & {86.8} \\
  &  & \multicolumn{2}{c|}{6:4 Averaging}  &{1607.8} & \textbf{324.6} & \textbf{76.5} & \textbf{76.3} & \textbf{65.9}(\textbf{73.6})&\textbf{43.7} & \textbf{38.8} & {82.3} & {64.5} & \textbf{80.4} & \textbf{87.4} \\
    \midrule
\multirow{3}{*}{Llama-3-8B} & \multirow{3}{*}{w$\backslash$} & B-695K + A-Ins & F  & {1599.6} & {348.9} & {76.3} & {75.6} & \textbf{66.2}({73.4}) & 41.6 & 38.3 & {81.9} & \textbf{64.1} & 79.5 & {86.4} \\
 &  & B-695K + L-665K + A-Ins & T  &\textbf{1649.7} & {341.1} & {77.2} & {76.5} & {65.5}({73.1})&{40.4} & 38.8 & {82.8} & \textbf{64.1} & {79.0} & \textbf{87.3} \\
  &  & \multicolumn{2}{c|}{4.2:5.8 Averaging (\textbf{Bunny-8B})}  &{1644.1} & \textbf{367.5} & \textbf{78.1} &\textbf{77.2} & \textbf{66.2}(\textbf{73.5})& \textbf{43.3} & \textbf{39.0} & \textbf{82.9} & {64.0} & \textbf{79.9} & {87.2} \\
\bottomrule
\end{tabular}}
    \label{tab:abl_merge}
\end{table}

%% file: tables/ablation1.tex
\begin{table}
    \centering
\caption{More results of Bunny with various language and vision backbones. The best performances are achieved by integrating SigLIP-SO \cite{zhai2023sigmoid} and Llama-3-8B \cite{llama3modelcard}.}
   \setlength{\tabcolsep}{3pt}

\resizebox{\textwidth}{!}{\begin{tabular}{ll|ccccccccccc}
\toprule
{Vision Encoder} & {LLM} & {MME$^\text{P}$} & {MME$^\text{C}$} &  MMB$^\text{T}$ & MMB$^\text{D}$ & {SEED(-I)} & MMMU$^\text{V}$ & MMMU$^\text{T}$ & {VQA$^\text{v2}$} & {GQA} & {SQA$^\text{I}$} & POPE \\
\midrule
\multirow{3}{*}{ EVA02-CLIP-L (0.4B) } & Phi-1.5 (1.3B)  & 1213.7 & 278.9 & 60.9 & 56.8 & 56.4(64.1) & 30.0 & 28.4 & 76.5 & 60.4 & 58.2 & 86.1 \\
& StableLM-2 (1.6B)  & 1301.0 & 235.0 & 58.4 & 56.4 & 55.3(62.8) & 29.8 & 29.4 & 74.6 & 56.7 & 60.0 & 84.8 \\
& Phi-2 (2.7B) & 1421.0 & 285.4 & 68.6 & 67.4 & 62.2(70.2) & 35.9 & 32.6 & 78.9 & 62.3 & 69.1 & 87.1 \\ \midrule
\multirow{5}{*}{ SigLIP-SO (0.4B) } & Phi-1.5 (1.3B) & 1230.0 & 237.5 & 61.2 & 59.7 & 57.7(65.3) & 30.0 & 29.1 & 78.0 & 61.1 & 61.3 & 85.8 \\
& StableLM-2 (1.6B)  & 1366.8 & 236.1 & 65.1 & 62.8 & 58.8(67.5) & 29.9 & 29.8 & 78.9 & 60.9 & 61.1 & 85.9 \\
& Phi-2 (2.7B) & 1488.8 & 289.3 & 69.2 & 68.6 & 62.5(70.7) & 38.2 & 33.0 & 79.8 & 62.5 & 70.9 & 86.8 \\
& Phi-3-Mini (3.8B) & {1581.5} & {361.1} & {75.7} & {74.2} & {64.9}({72.5}) & {41.4} & {38.4} & {82.1} & 63.2 & {78.3} & {87.2} \\
& Llama-3-8B & {1644.1} & {367.5} & {78.1} & {77.2} & {66.2}({73.5})& {43.3} & {39.0} & {82.9} & {64.0} & {79.9} & {87.2}\\
\bottomrule
\end{tabular}}
    \label{tab:abl1}
\end{table}



%% file: figures/demo_table.tex
\newcommand\HumanIcon{\raisebox{0.3ex}{\tikz\fill[scale=0.4] (0,0) circle (1ex);}}
\newcommand\BunnyIcon{\tikz\draw[scale=0.4,circle,fill=gray] (0,0) circle (1ex);}
\begin{table}[h]
 \small
  \centering
  \begin{tabular}{  m{3cm} | m{10cm}  }
    \hline
    \begin{minipage}[t]{\linewidth}
		\centering
		\includegraphics[width=0.9\linewidth]{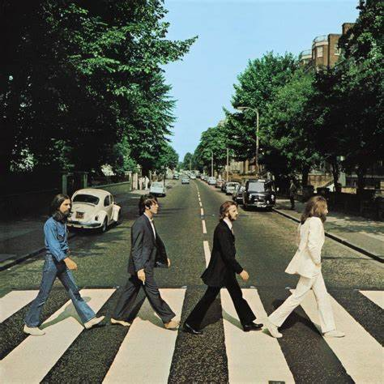}
	\end{minipage}
    & \vspace{0.5ex} {\HumanIcon} \textbf{Do you know who are they?}\newline They are members of the Beatles, one of the most famous rock bands in history.\newline
    {\HumanIcon} \textbf{Do you recognize where does this picture come from?}\newline Yes, this is a famous photo taken by Iain Macmillan on Abbey Road in London, England. It was used as the cover art for the Beatles' album ``Abbey Road'' released in 1969.\newline
    {\HumanIcon} \textbf{How do you know?}\newline The image has been widely recognized due to its iconic status and the distinctive white lines that cross the street, which are part of the pedestrian crossing at Abbey Road. \vspace{0.5ex} \\ \hline
        
    \begin{minipage}[t]{\linewidth}
        \centering
        \includegraphics[width=0.9\linewidth]{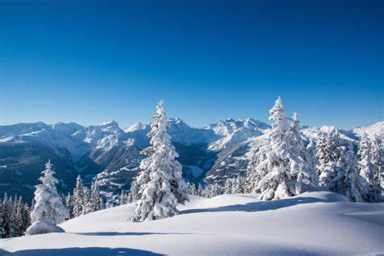}
    \end{minipage}
    & \vspace{0.5ex} {\HumanIcon} \textbf{What kind of sport can I play here?}\newline Given the snowy environment, you could engage in winter sports such as skiing or snowboarding.\newline
    {\HumanIcon} \textbf{Is it safe for me to go skiing here?}\newline It appears to be a suitable location for skiing due to the presence of snow-covered slopes. However, it would be best to check with local authorities for any safety guidelines or restrictions before engaging in this activity.  \vspace{0.5ex} \\ \hline
    
    \begin{minipage}[t]{\linewidth}
        \centering
        \vspace{-10ex} 
        \includegraphics[width=0.9\linewidth]{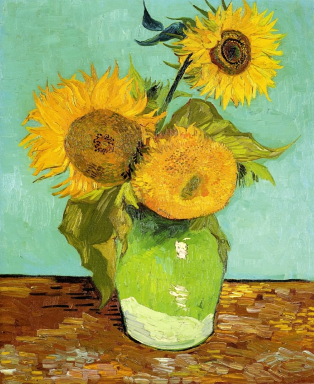}
        \vspace{1ex} 
    \end{minipage}
    & \vspace{4ex} {\HumanIcon} \textbf{What kind of art is this?}\newline This is a painting by Vincent van Gogh, specifically a still life featuring sunflowers in a vase.\newline
    {\HumanIcon} \textbf{Do you find the painting beautiful?}\newline As an AI, I don't have personal opinions or emotions. However, many people consider Van Gogh's paintings to be beautiful due to their vibrant colors and unique style. \\ \hline
    
    \begin{minipage}[t]{\linewidth}
        \centering
        \includegraphics[width=0.9\linewidth]{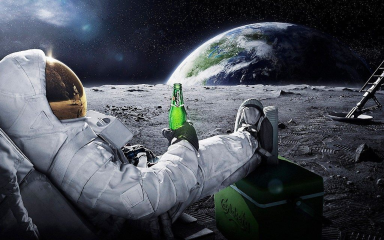}
    \end{minipage}
    & \vspace{0.5ex} {\HumanIcon} \textbf{What is the astronaut holding in his hand?}\newline The astronaut is holding a green bottle of beer in his hand.\newline
    {\HumanIcon} \textbf{Can this happen in real world?}\newline It's highly unlikely that an astronaut would be sitting on the moon with a beer bottle. This image seems to be a creative representation or a humorous depiction rather than a realistic scenario.\newline
    {\HumanIcon} \textbf{Why is that?}\newline Astronauts are trained for space missions and their primary focus is on scientific research and exploration. They follow strict protocols and guidelines to ensure safety and efficiency during their time in space. Drinking beer on the moon would not align with these objectives. \vspace{0.5ex} \\ \hline
    
    \begin{minipage}[t]{\linewidth}
        \centering
        \includegraphics[width=0.9\linewidth]{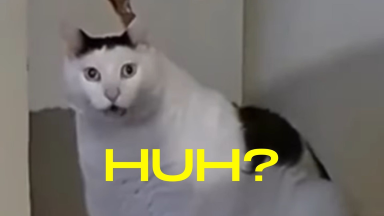}
    \end{minipage}
    & \vspace{0.5ex} {\HumanIcon} \textbf{Why is the image funny?}\newline The image is funny because it features a cat sitting in front of a wall with yellow letters spelling out ``Huh?'' on it. This creates a humorous juxtaposition between the cat's expression and the question mark, suggesting that the cat might be confused or puzzled by something.\newline
    {\HumanIcon} \textbf{When should I use the meme?}\newline You can use this meme when you want to express confusion or puzzlement about something. It could also be used humorously to ask for clarification or to poke fun at someone who seems perplexed. \vspace{0.5ex} \\ \hline
    
    \begin{minipage}[t]{\linewidth}
        \centering
        \includegraphics[width=0.9\linewidth]{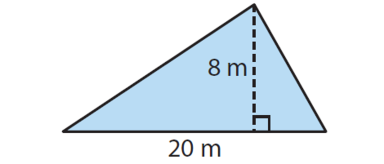}
    \end{minipage}
    & \vspace{0.5ex} {\HumanIcon} \textbf{What is the height and base of the triangle?}\newline The height of the triangle is 8 meters and its base is 20 meters.\newline
    {\HumanIcon} \textbf{Can you calculate the area of triangle?}\newline Yes, I can calculate the area of the triangle. The formula for the area of a triangle is (base * height) / 2. In this case, it would be (20m * 8m) / 2 = 80 square meters.\newline
    {\HumanIcon} \textbf{Students of what grade should be able to solve this problem?}\newline This problem should be solvable by students in middle school or high school who have learned basic geometry concepts such as calculating the area of triangles using the formula (base * height) / 2. \vspace{0.5ex} \\ \hline
  \end{tabular}
  \vspace{2pt}
  \caption{\textbf{Qualitative performances of Bunny-8B}: The bold text represents user input, while the non-bold text is the model response. The results demonstrate Bunny's versatile capabilities in visual understanding, common sense reasoning, external knowledge referring, optical character recognizing, and mathematical problem solving.}
\label{fig:demo}
\end{table}

%% file: figures/demo_table_zh.tex
\begin{CJK*}{UTF8}{gbsn}
\begin{table}[h]
 \small
  \centering
  \begin{tabular}{  m{3cm} | m{10cm}  }
    \hline
    \begin{minipage}[t]{\linewidth}
		\centering
		\includegraphics[width=0.9\linewidth]{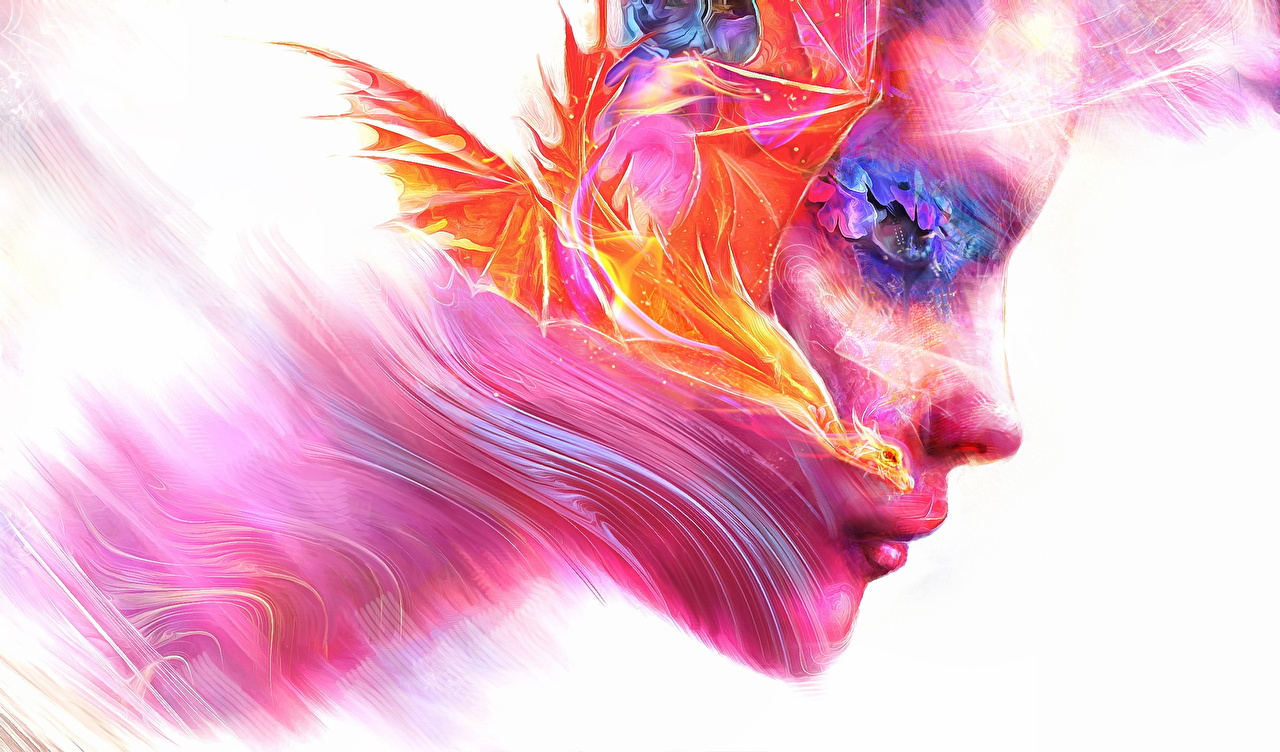}
	\end{minipage}
    & \vspace{1ex} {\HumanIcon} \textbf{请用中文鉴赏一下图片？}\newline 这是一幅奇幻的绘画，描绘了一位女性和一只红色的龙。女性的脸部被细腻地描绘，包括她的眼睛、鼻子、嘴巴和耳朵。她长着一头卷曲的发型，发色是粉红色的。同时，她的皮肤也呈现出一种粉红色的光泽。\newline 在女性面前飞舞着一只红色的龙，龙的身体部分可以看到它的翅膀和尾巴。龙的颜色非常鲜艳，体现了绘画中奇幻元素的特点。\newline 整体来说，这幅画充满了奇幻和梦幻的氛围，女性和龙之间的互动关系也很有趣。 \vspace{1ex} \\ \hline

    \begin{minipage}[t]{\linewidth}
        \centering
        \includegraphics[width=0.9\linewidth]{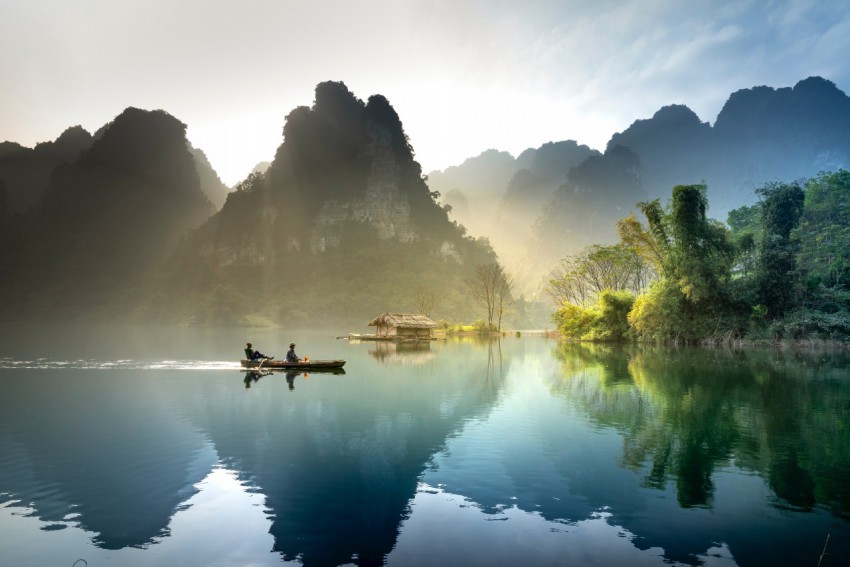}
    \end{minipage}
    & \vspace{1ex} {\HumanIcon} \textbf{请用中文详细描述这张图片}\newline 这是一幅描绘了一个美丽的自然景观的画面。画面中，两个人在一艘小船上，正在穿过一片水面。这艘船位于画面的中心，人们似乎是在享受着这个宁静的湖泊的美景。\newline 水面上还有一些反射，可能是来自周围的山脉和树林的光芒。这些反射使得水面更加迷人，增添了一种神秘感。\newline 在画面背景中，有一座山脉，它们高耸入云，形成了了一个壮丽的天际线。山脉的山峰部分被阳光照亮，反映在水面上，给整个画面带来了温暖的氛围。\newline 在山脉的脚下，是一片茂密的树林。这些树林的叶子呈现出不同的颜色，可能是由于季节的变化或日落的光芒。这些树林的反影也可以在水面上看到，这使得整个画面更加生动。\newline 总体来说，这幅画面展示了一个宁静、美丽的自然景观，两个人在其中度过了一段愉快的时光。 \vspace{1ex} \\ \hline

    \begin{minipage}[t]{\linewidth}
        \centering
        \includegraphics[width=0.9\linewidth]{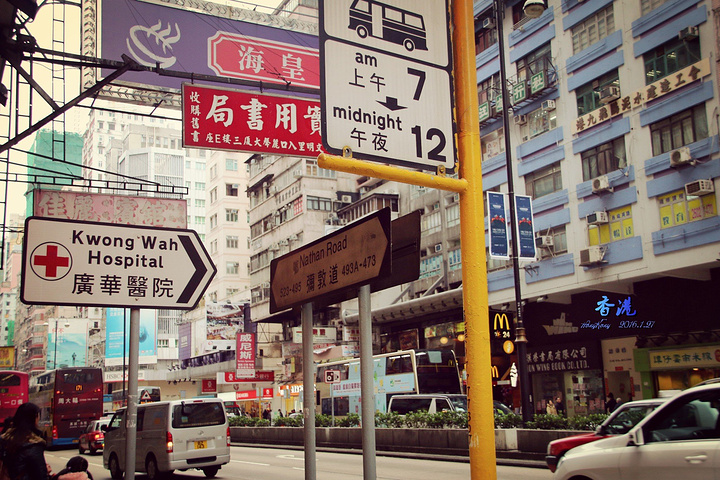}
    \end{minipage}
    & \vspace{1ex} {\HumanIcon} \textbf{路牌上的医院名称是什么？}\newline 路牌上的医院名称是``Kwong Wah Hospital''。\newline
    {\HumanIcon} \textbf{巴士的运营时间是几点？}\newline 巴士的运营时间是从早上7点到晚上12点。\newline
    {\HumanIcon} \textbf{这是哪座城市？}\newline 这是香港的市区。 \vspace{1ex} \\ \hline

    \begin{minipage}[t]{\linewidth}
        \centering
        \vspace{-7ex} 
        \includegraphics[width=0.9\linewidth]{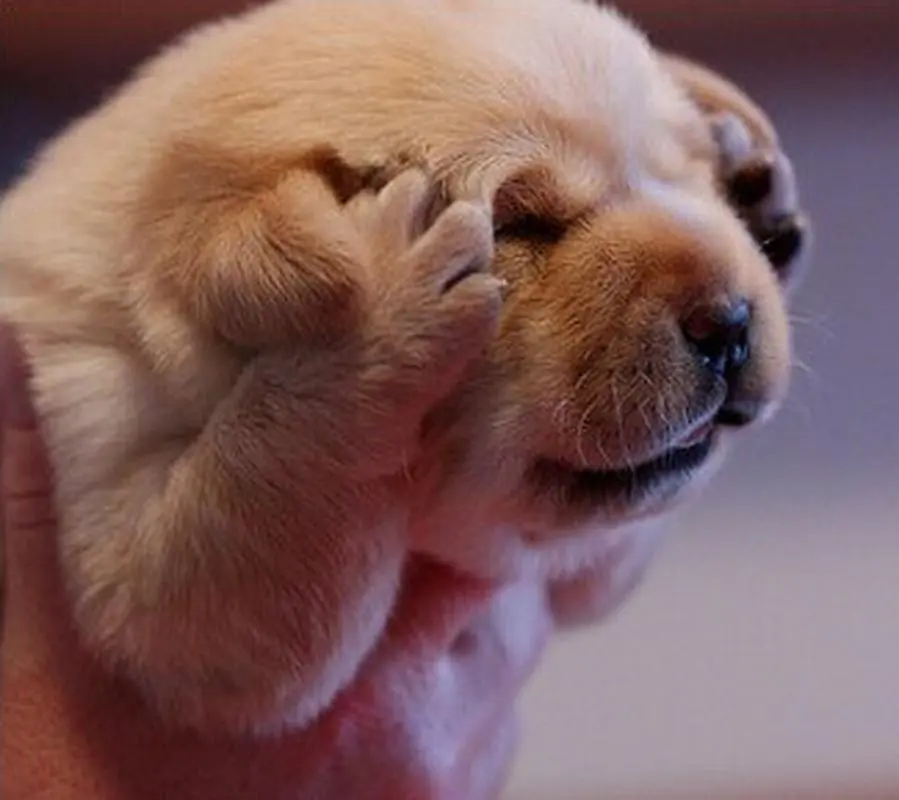}
        \vspace{1ex} 
    \end{minipage}
    & \vspace{3ex} {\HumanIcon} \textbf{请用中文回答为什么这张图片如此有趣？}\newline 这张图片非常有趣，因为它展示了一个小狗的可爱和天真。小狗的双手抱着自己的头，表明它在思考或感到困惑。这场景让人联想到人类的行为，特别是孩子们经常这样做。同时，小狗的颜色和表情也很吸引人，让人无法不被其可爱所吸引。 \\ \hline

    \begin{minipage}[t]{\linewidth}
        \centering
        \vspace{-10ex} 
        \includegraphics[width=0.9\linewidth]{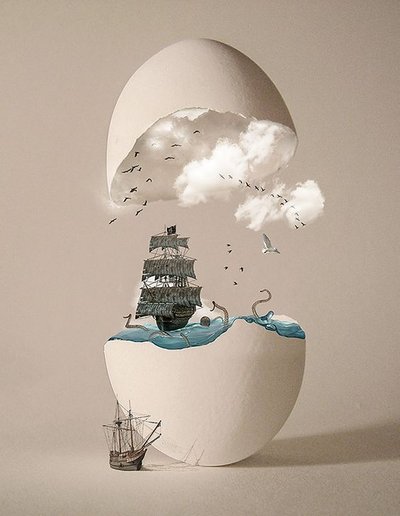}
        \vspace{1ex} 
    \end{minipage}
    & \vspace{6ex} {\HumanIcon} \textbf{这张图的创意体现在什么地方？}\newline 这张图的创意体现在将一个船和一只鸟安置在一个蛋中，创造出一个奇怪、有趣的场景。\newline
    {\HumanIcon} \textbf{这张图片可能具有什么用途？}\newline 这张图可能是用于表达创意或设计的概念，或者作为一种艺术作品展现。 \\ \hline
    
  \end{tabular}
  \vspace{2pt}
  \caption{\textbf{Qualitative performances of Bunny-8B}: The results demonstrate its good Chinese instruction-following ability.}
\label{fig:demo_zh}
\end{table}
\end{CJK*}